\pdfoutput=1 %
\PassOptionsToClass{arxiv}{tvst-manuscript}
\documentclass[review]{tvst-manuscript}

\graphicspath{{figures/}}
\usepackage{colortbl}%
\newcommand{\resulttablescale}{0.79}%

\newcommand{\SoftwareName}{AA-OCTSeg}%

\abstractwordlimit{250} %

\title{\SoftwareName: An Anatomy-Aware Framework for Cross-Device Optical Coherence Tomography Segmentation}

\author[1]{Yang Zhang}
\author[1]{Rong Zhou}
\author[1]{Qingyu Chen}
\affil[1]{Department of Biomedical Informatics and Data Science, Yale School of Medicine, New Haven, CT, USA}

\wordcount{\placeholder{XXXX}}

\disclosure{Y. Zhang, None; R. Zhou, None; Q. Chen, None}

\keywords{optical coherence tomography, domain adaptation, retinal layer segmentation, semi-supervised learning, artificial intelligence in ophthalmology}

\begin{document}

\maketitle

\begin{tvstabstract}
  \Purpose{Retinal layer segmentation in optical coherence tomography (OCT) is essential for quantitative assessment of ocular anatomy and disease but remains challenging across imaging devices and complex retinal anatomies. To address this, we developed \SoftwareName{} (Anatomy-Aware OCT Segmentation), an open-source framework for cross-device 2D OCT segmentation.}
  \Methods{\SoftwareName{} incorporates anatomy-aware augmentations to model variations in retinal morphology and anatomy-aware perturbations to simulate local signal degradation. Semi-supervised cross-teaching enables a SAM-based network and a U-Net to learn from labeled and unlabeled scans from the AI-READI, JHU HC-MS, Duke DME, and OCT5k datasets. Predictions are fused and refined using Viterbi decoding to enforce anatomical layer ordering.}
  \Results{\SoftwareName{} ranked first in the MICCAI 2026 Domain Adaptation for OCT Segmentation Challenge, achieving a score of 0.8068 on the hidden test set. On an external Duke AMD dataset from an unseen Bioptigen device, \SoftwareName{} achieved the lowest mean absolute error of 2.49 pixels among the evaluated methods and a Dice score of 84.0\%. Ablation studies showed that anatomy-aware augmentation and anatomical post-processing improved robustness.}
  \Conclusions{Cross-device OCT segmentation is particularly challenging around the fovea, optic disc, and retinal lesions. Incorporating anatomical priors throughout augmentation, semi-supervised learning, and post-processing improves segmentation robustness across devices and scan types while reducing reliance on manual annotations.}
  \TranslationalRelevance{\SoftwareName{} provides an open-source framework and trained models for retinal layer segmentation across OCT devices, supporting research and clinical applications that rely on quantitative retinal measurements. Our code is available at \url{https://github.com/Yale-BIDS-Chen-Lab/yzjaxon-DA-OCT-Challenge-MICCAI-2026}.}
\end{tvstabstract}
\section{Introduction}
Optical coherence tomography (OCT) is a non-invasive imaging technique in ophthalmology that can capture high-resolution, cross-sectional views of retinal anatomical and pathological structures.  Since its invention in 1991, OCT has become a widely adopted technology using low-coherence near-infrared interferometry.\cite{huang1991oct,fujimoto2016oct} Retinal layer segmentation remains one of the fundamental tasks when applying OCT to assist diagnosis. The morphological features of specific layers serve as critical biomarkers for diagnosing and monitoring ocular diseases.\cite{bussel2014glaucoma,petzold2017ms,schmidterfurth2016biomarkers} For example, retinal thickness variation is a crucial indicator for macular edema.\cite{hee1995edema,browning2007thickness} While most commercial OCT devices offer automatic retinal layer segmentation, they output a limited number of distinguishable layers and are highly susceptible to failure in the presence of severe speckle noise or disruptive retinal lesions.\cite{heikka2020noise,deazevedo2020manual} Consequently, developing robust, automated OCT segmentation algorithms remains a highly active area of research.\cite{zhang2025review}

Driven by remarkable advances in artificial intelligence (AI),\cite{zhou2023retfound} deep learning (DL) approaches have achieved state-of-the-art performance in retinal layer segmentation.\cite{roy2017relaynet,tan2024tcct,morano2025mirage,fazekas2026sdretinanet} Recent foundation models have further expanded the use of large-scale retinal imaging data for transferable ophthalmic representations.\cite{zhou2023retfound,chang2026ophmae,hou2026natural} Despite their great success, several crucial gaps remain that limit their clinical utility. One of the most essential ones is the difficulty in cross-device generalization. Commercial OCT devices differ substantially in their hardware optics, scanning protocols, and post-processing pipelines, resulting in significant distributional shifts across image datasets.\cite{romobucheli2020variability} A DL model trained on one device can easily fall short when processing images from a different device, or when the required number of layers differs from that of its training data.\cite{romobucheli2020variability,wu2023crossdomain,he2020sdanet} Furthermore, conventional supervised DL models demand extensive, pixel-level annotations from clinical experts.\cite{tajbakhsh2020imperfect} Collecting full annotations for every vendor requires substantial effort and costs. Labeled data across different devices remains limited. Together, these issues pose a common challenge: how to utilize limited labeled data from several devices to train an effective model that can generalize well to unseen devices?

Recent studies have explored domain adaptation and semi-supervised learning to improve the generalizability of retinal OCT segmentation across devices.\cite{he2020sdanet,koch2022noise,chen2023harmonization,lu2025ssda,sedai2019uncertainty,liu2019semi,fazekas2022sdlayernet,lu2023boundary,sreng2025widefield,fazekas2026sdretinanet,fan2026dplseg,li2026dualcurriculum} However, most existing approaches primarily address differences in image appearance or are developed for standard macular scans, which cover only a limited central region of the retina. In practice, cross-device variation is compounded by substantial anatomical and scan-level heterogeneity. Widefield OCT can span the fovea and optic nerve head (ONH),\cite{hood2016widefield,chaglasian2018maestro} where retinal curvature, layer thickness, and boundary continuity differ markedly from those observed in conventional macular scans.\cite{devalla2018drunet,zang2019peripapillary,li2021peripapillary} Pathological changes can further distort normal retinal anatomy and obscure layer interfaces. These variations create failure modes that cannot be fully addressed by appearance-level domain adaptation alone. Existing approaches also tend to consider cross-device adaptation, semi-supervised learning, and anatomical priors separately,\cite{chen2023harmonization,sedai2019uncertainty,he2021structured,liu2024alignment} motivating a unified framework that explicitly incorporates retinal anatomy throughout training and inference.

To address these challenges, we propose Anatomy-Aware OCT Segmentation (\SoftwareName), a unified framework for cross-device retinal OCT segmentation that incorporates anatomical priors throughout the learning and inference pipeline. Specifically, \SoftwareName{} introduces anatomy-aware augmentations to model variations in retinal curvature, layer thickness, and optic-disc morphology, together with anatomy-aware perturbations that simulate localized signal degradation. To leverage limited labeled data across devices, we further employ a semi-supervised cross-teaching strategy between two complementary segmentation networks using both labeled and unlabeled OCT scans. At inference, the predictions are fused and refined using anatomically constrained Viterbi decoding to enforce the expected ordering of retinal layers. \SoftwareName{} was trained using the recently released AI-READI dataset\cite{aireadi2024rethinking,aireadi2025flagship} and multiple public OCT datasets\cite{he2019jhu,chiu2015dme,arikan2025oct5k} and achieved first place in the MICCAI 2026 Domain Adaptation for OCT Segmentation Challenge.\cite{owen2026daoct}

\section{Methods}
\begin{figure}[htbp]
  \centering
  \includegraphics[width=\linewidth,
    alt={Diagram of the workflow in three panels: augmentation, training in a
    supervised and a semi-supervised stage, and inference}]{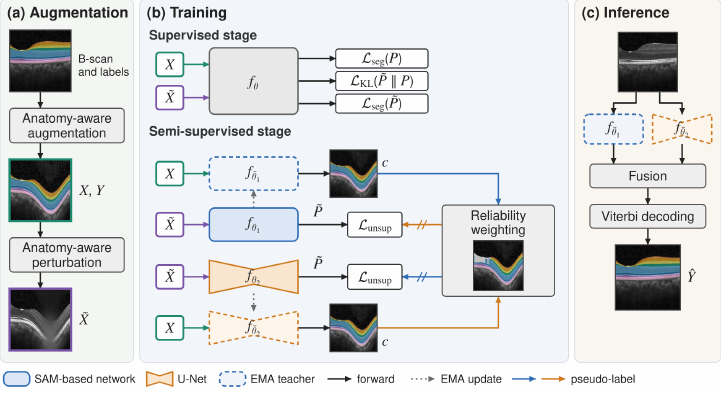}
  \caption{Overview of the \SoftwareName{} workflow. (a) Anatomy-aware
    augmentation and perturbation produce a training patch $X$ with labels
    $Y$ and its perturbed view $\tilde X$. (b) In the supervised stage, each
    network is trained on both views. In the semi-supervised stage, the
    teacher of each network generates pseudo-labels $c$ for the student of
    the other network. (c) At inference, the
    predictions of the two networks are fused and decoded with the Viterbi algorithm.}
  \label{fig:workflow}
\end{figure}

Training of \SoftwareName{} consists of a supervised stage followed by a
semi-supervised stage. We first train two segmentation networks independently
on fully and partially annotated OCT B-scans. We then utilize semi-supervised
cross-teaching that allows each network's teacher to generate pseudo-labels for
the other network to learn on unlabeled scans. In both stages, we use anatomy-aware augmentations to model variations in retinal shape and anatomy-aware perturbations to simulate local signal degradation.
Figure~\ref{fig:workflow} gives an overview of the workflow.

\subsection{Preliminaries}
Given a normalized OCT B-scan $X\in\mathbb{R}^{H\times W}$, our goal is to
assign one of ten classes to each pixel. We denote the underlying class label
map by $Y\in\{0,\ldots,9\}^{H\times W}$. At pixel $(i,j)$, $X_{ij}$ is the
image intensity and $Y_{ij}$ is the class label, with $i$ indexing depth and
$j$ indexing lateral position. Each segmentation network $f_\theta$ outputs a
pixelwise probability map $P=f_\theta(X)\in[0,1]^{H\times W\times 10}$, where
$P_{ijc}$ is the predicted probability that $Y_{ij}=c$.

Let $b_k(j)$ denote the row coordinate of the $k$-th retinal interface in
column $j$. The nine interfaces follow a fixed anatomical order from shallow to
deep, so $b_1(j)\le\cdots\le b_9(j)$ in each column. In other words, the class label is nondecreasing with depth.

\paragraph{Partial annotations.} Some training datasets provide annotations for
only a subset of the nine interfaces. In these scans, we use the available
interfaces in each column to determine which classes are admissible at each
pixel. An annotated interface at or above pixel $(i,j)$, $b_k(j)\le i$, implies
$Y_{ij}\ge k$. An annotated interface below the pixel, $b_k(j)>i$, implies
$Y_{ij}\le k-1$. We denote the largest of these lower bounds by $L_{ij}$ and the smallest of
these upper bounds by $U_{ij}$. When the corresponding bound is unavailable, we set $L_{ij}=0$ and $U_{ij}=9$. The integers from $L_{ij}$ to $U_{ij}$ are
the admissible classes for the pixel. When $L_{ij}=U_{ij}$, the pixel has an
exact label. We denote by $\Omega$ the set of pixels in columns with at least
one annotated interface.

\subsection{Anatomy-Aware Augmentation}
Retinal curvature and layer thickness vary across OCT B-scans. Scans that include the optic disc often show a substantial local depression. To model these variations, we introduce two anatomy-aware augmentations. Local morphological deformation models local changes in retinal curvature and thickness, while the optic-disc-like depression mimics the drastic structural changes near the optic disc. Both augmentations shift pixels along the depth axis within each column and vary the displacement smoothly across columns.

Let $d(i,j)$ denote the signed axial displacement at pixel $(i,j)$, with
positive values indicating a downward shift. A pixel at depth $i$ in column $j$
moves to a new depth
\[
  m_j(i)=i+d(i,j).
\]
We require $m_j$ to be strictly increasing with depth so that the retinal
interfaces retain their anatomical order. The warped position of interface $k$
is $b_k^{\mathrm{warp}}(j)=m_j(b_k(j))$, and its axial displacement is
\[
  \Delta b_k(j)=b_k^{\mathrm{warp}}(j)-b_k(j)=d\!\left(b_k(j),j\right).
\]

\subsubsection{Local Morphological Deformation}
To mimic local variations in retinal curvature and layer thickness, we use two
types of deformation. An axial shift bends the retina by moving all layers
together within each column, while a thickness modulation expands or compresses
layers by varying the displacement with depth. For each augmented B-scan, we
combine $N$ such local deformations into a single displacement field,
$d(i,j)=\sum_{n=1}^{N}d_n(i,j)$, where $N$ is randomly chosen to be one or two.
Each deformation has an amplitude $A_n$ with a randomly chosen sign. The
displacement magnitude follows a Gaussian profile centered at column $j_n$,
with a full width at half maximum of 10--30\% of the image width $W$.

\paragraph{Axial shift.} An axial shift moves all depths within a column by the
same amount,
\[
  d_n(i,j)
  =
  A_n\exp\!\left[-\frac{(j-j_n)^2}{2\sigma_{\mathrm{lat},n}^{2}}\right],
\]
where $\sigma_{\mathrm{lat},n}$ controls the lateral extent. As the shift
varies across columns, it bends the retina while preserving the thickness of
each layer. This models the locally varying curvature seen in widefield scans.

\paragraph{Thickness modulation.} A thickness modulation multiplies the same
displacement by a Gaussian depth weight, giving
\[
  d_n(i,j)
  =
  A_n\exp\!\left[-\frac{(j-j_n)^2}{2\sigma_{\mathrm{lat},n}^{2}}\right]
  \exp\!\left[-\frac{(i-i_n)^2}{2\sigma_{\mathrm{ax},n}^{2}}\right].
\]
The displacement magnitude is greatest at depth $i_n$ and decreases with
distance from it over an axial scale $\sigma_{\mathrm{ax},n}$. This expands
layers on one side of $i_n$ and compresses those on the other, producing local
thickness changes such as those around the foveal pit. To preserve depth order and limit local axial stretching, we multiply the summed displacement field by a single factor in $(0,1]$ when necessary. For the modulation to act on the retinal layers, $i_n$ must lie near the retina. Since retinal layers appear as a bright band in OCT scans, we estimate the depth of the retina as the mean row coordinate of pixels whose intensity exceeds the image mean by half a standard deviation, and set $i_n$ to this depth plus a random axial offset.

\subsubsection{Optic-disc-like Depression}
The second augmentation mimics the abrupt depression of the retinal layers near the optic disc. Within each column, we move the inner limiting membrane ($b_1$) farther downward than Bruch's membrane ($b_9$). The difference between their displacements determines the degree of axial compression applied to the intermediate retinal layers. Across columns, the nominal downward displacement decreases gradually from the center
of the depression to zero at its lateral edges.

In column $j$, the inner limiting membrane moves downward by
$\Delta b_1(j)\ge 0$ pixels. Bruch's membrane moves downward by a fraction
$r_{\mathrm{BM}}$ of this distance, so that
$\Delta b_9(j)=r_{\mathrm{BM}}\Delta b_1(j)$, with $0<r_{\mathrm{BM}}<1$. Their
separation after warping is therefore
\[
  b_9^{\mathrm{warp}}(j)-b_1^{\mathrm{warp}}(j)
  =
  [b_9(j)-b_1(j)]-(1-r_{\mathrm{BM}})\,\Delta b_1(j).
\]

Tissue above $b_1$ moves with it by $\Delta b_1(j)$. Below $b_9$, the
displacement decreases linearly from $\Delta b_9(j)$ at its original depth to
zero at the fixed row $b_9^{\mathrm{warp}}(j)+\ell$, which lies $\ell>0$ pixels
below the displaced interface. All deeper rows have zero displacement. Between
$b_1$ and $b_9$, we linearly interpolate the displacement with depth. For each
intermediate interface $b_k$, with $k=2,\ldots,8$, this gives
\[
  \Delta b_k(j)
  =
  \Delta b_1(j)
  +
  \frac{b_k(j)-b_1(j)}{b_9(j)-b_1(j)}
  \bigl(\Delta b_9(j)-\Delta b_1(j)\bigr).
\]
Each interface therefore retains its relative position between $b_1$ and $b_9$, and each intermediate layer retains the same fraction of its original thickness.

Having defined how the other interfaces move for a given shift of $b_1$, we now
determine that shift for each column, starting at the center of the depression.
At the center column $j_{\mathrm c}$, we compute the shift needed to retain a
target fraction $r_{\mathrm{keep}}$ of the original $b_1$--$b_9$ separation by
setting $b_9^{\mathrm{warp}}(j_{\mathrm c})-b_1^{\mathrm{warp}}(j_{\mathrm c})
=r_{\mathrm{keep}}[b_9(j_{\mathrm c})-b_1(j_{\mathrm c})]$ in the separation
equation above and solving for $\Delta b_1(j_{\mathrm c})$. Capping this shift
at $A_{\max}$ gives the nominal center amplitude
\[
  A_{\mathrm c}
  =
  \min\!\left\{
  A_{\max},\,
  \frac{(1-r_{\mathrm{keep}})\,[b_9(j_{\mathrm c})-b_1(j_{\mathrm c})]}
       {1-r_{\mathrm{BM}}}
  \right\}.
\]

To make the depression shallower toward either edge, we scale the center shift
$A_{\mathrm c}$ at column $j$ by a lateral weight $w_{\mathrm{disc}}(j)$. The
lateral weight equals 1 at $j_{\mathrm c}$ and decreases smoothly to 0 over
$h_L$ columns to the left and $h_R$ columns to the right. We refer to the
columns $j_{\mathrm c}-h_L\le j\le j_{\mathrm c}+h_R$ as the depression window.
Within this window, we define the weight as
\[
  w_{\mathrm{disc}}(j)
  =
  \frac{1+\cos\!\left(\pi |j-j_{\mathrm c}|/h\right)}{2},
\]
where $h=h_L$ for $j\le j_{\mathrm c}$ and $h=h_R$ for $j>j_{\mathrm c}$. When
$h_L$ and $h_R$ differ, the weight $w_{\mathrm{disc}}(j)$ decreases from 1 to 0
over more columns on the wider side.

For each column $j$ within the window, we first compute a downward shift of
$A_{\mathrm c}w_{\mathrm{disc}}(j)$ for $b_1$. Since $b_9$ moves down by
$r_{\mathrm{BM}}\Delta b_1(j)$, the distance between the two interfaces
decreases by $(1-r_{\mathrm{BM}})\Delta b_1(j)$. To keep them at least
$t_{\min}$ pixels apart after warping, we limit the shift of $b_1$ to
\[
  \Delta b_1(j)
  =
  \min\!\left\{
  A_{\mathrm c}w_{\mathrm{disc}}(j),\,
  \frac{b_9(j)-b_1(j)-t_{\min}}{1-r_{\mathrm{BM}}}
  \right\}.
\]
For columns outside the window, we set $w_{\mathrm{disc}}(j)=\Delta b_1(j)=0$.
With $t_{\min}>0$ and $\ell>0$, the depth map is strictly increasing and
requires no further scaling.

We apply this augmentation only where both interfaces are annotated and at
least 16 pixels apart throughout the window. The entire window must lie at
least $0.08W$ from the column where $b_1$ is deepest, which we use as a
landmark for the foveal center in macular scans. We exclude scans acquired with
an optic disc protocol from this augmentation.

After transformation, we darken the tissue between the displaced interfaces. For
pixels of $X^{\mathrm{warp}}$ with
$b_1^{\mathrm{warp}}(j)\le i<b_9^{\mathrm{warp}}(j)$, we blend the warped
intensity with the 10th-percentile intensity of the warped image, using a
weight proportional to $w_{\mathrm{disc}}(j)$.
We fill the rows vacated at the top of the image with Gaussian noise matched to
the intensity statistics immediately above $b_1$.

\subsection{Anatomy-Aware Perturbation}
The anatomy-aware augmentations described above model variations in retinal shape. In
practice, retinal interfaces are often partly obscured by vessel shadows, media
opacities, and local losses of focus or signal \cite{chhablani2014artifacts,bazvand2020artifacts}. These degradations typically
affect a band of adjacent columns. To encourage the models to infer an obscured
interface from the surrounding visible structure, we  perturb the intensities
within a randomly placed band of columns. Anatomy-aware perturbation produces a second view of
the same patch. We use $X$ to denote the normalized training patch after the
axial and standard augmentations and $\tilde X$ to denote its spatially
aligned, perturbed view.

As in the optic-disc-like depression, a lateral weight $w_{\mathrm{pert}}(j)$
controls the strength of the perturbation in column $j$. It follows a Gaussian profile,
\[
  w_{\mathrm{pert}}(j)
  =
  \exp\!\left[-\frac{(j-j_0)^2}{2\sigma_{\mathrm{lat}}^2}\right],
\]
where the center column $j_0$ is randomly chosen. The spread
$\sigma_{\mathrm{lat}}$ is chosen so that the profile has a full width at half
maximum of 10--30\% of the patch width. The weight equals 1 at $j_0$ and
decreases smoothly on both sides, so the perturbation gradually fades with distance from the center.

To generate the perturbed view, we initialize $\tilde X$ as a copy of $X$. For each sample, we randomly select one or two of the following five perturbations and
apply them in sequence within the same band.

\paragraph{Blur.} We blend the current image with its Gaussian-smoothed
version.
This attenuates fine layer texture within the band and mimics a local loss of
focus.

\paragraph{Noise.} We add Gaussian noise whose amplitude follows the lateral
weight.
We set the noise amplitude relative to the
intensity standard deviation of the unperturbed patch. This
decreases the signal-to-noise ratio within the band.

\paragraph{Contrast reduction.} We blend the current image with its local mean,
which we obtain using a wide Gaussian filter.
This reduces intensity differences between adjacent layers.

\paragraph{Shadow.} We attenuate the signal below an onset depth
$i_{\mathrm{sh}}$ to mimic shadowing beneath a vessel or an opacity. We use
the 5th-percentile intensity of the unperturbed patch $X$ as the
background level and scale the signal above background by a factor that
decreases smoothly with depth around $i_{\mathrm{sh}}$.
The scaling factor is close to 1 outside the shadowed region and approaches
its minimum at the band center well below $i_{\mathrm{sh}}$.

\paragraph{Local patch.} We add a bright or dark patch that is also localized
in depth. This
models a local bright artifact or a local loss of signal.

\begin{figure}[htbp]
  \centering
  \includegraphics[width=\linewidth,
    alt={Example B-scans before and after the anatomy-aware augmentations,
    with their displacement fields, and before and after the five
    perturbations}]{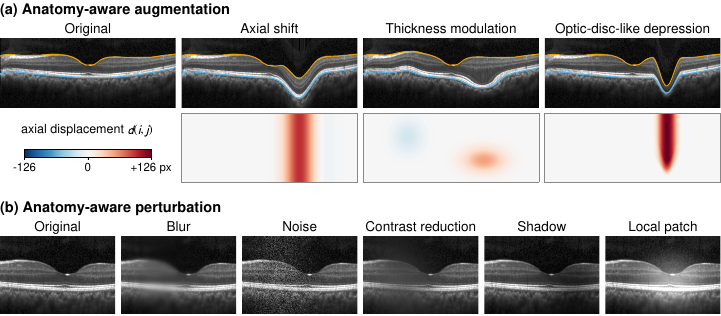}
  \caption{Examples of anatomy-aware augmentation and perturbation. (a) A
    B-scan before and after an axial shift, a thickness modulation, and the
    optic-disc-like depression, with the axial displacement $d(i,j)$ of each
    shown below. The orange and blue curves mark the inner limiting membrane
    ($b_1$) and Bruch's membrane ($b_9$). (b) A patch and its perturbed views
    under each of the five perturbations.}
  \label{fig:augmentation}
\end{figure}

We place $i_{\mathrm{sh}}$ and the center of the local patch near the retina using the same
procedure as for $i_n$ by taking the mean row coordinate of the bright pixels
in $X$ and adding a random axial offset. The retinal interfaces $b_k(j)$ have the same coordinates in $X$ and $\tilde X$. Figure~\ref{fig:augmentation} shows examples of the anatomy-aware augmentations and perturbations.

\subsection{Anatomy-Guided Semi-Supervised Learning}
Supervised learning is commonly used for OCT segmentation. However, it relies
on large annotated datasets that require substantial time and expertise to
collect. Large unlabeled cohorts often include scans acquired with diverse
protocols, such as widefield and optic-disc-centered scans. To make use of
these scans, we adopt a semi-supervised approach in which two networks provide
pseudo-labels for one another. We weight these pseudo-labels according to the
reliability of the retinal interfaces bounding each predicted region.

\subsubsection{Network Architectures}
Our framework incorporates two segmentation networks with different
architectures. The first network, $f_{\theta_1}$, combines a pre-trained image
encoder from the Segment Anything Model \cite{kirillov2023sam} with a multi-scale
convolutional decoder. The second network, $f_{\theta_2}$, is a convolutional
U-Net \cite{ronneberger2015unet} trained from scratch. At inference, the predictions
from both networks are fused to produce the final segmentation.

\subsubsection{Cross-Teaching}
For each network $s\in\{1,2\}$, we maintain a student with parameters
$\theta_s$ and a teacher with parameters $\bar\theta_s$. After each gradient
update of the student, we update the teacher parameters using an exponential
moving average.
Averaging the student parameters over training
steps helps stabilize the teacher's predictions \cite{tarvainen2017meanteacher}.

Let $X$ be an unlabeled OCT patch and $\tilde X$ its spatially aligned,
perturbed view. Each teacher generates pseudo-labels from $X$. Each student
predicts on $\tilde X$ and is trained to reproduce the pseudo-labels generated
by the other network's teacher.

\paragraph{Reliability weighting.} To limit the influence of erroneous
pseudo-labels and reduce confirmation bias, we assess the reliability of each
predicted interface. Below, we describe the direction in which teacher 2
supervises student 1. Let $Q=f_{\bar\theta_2}(X)$ denote the class
probabilities predicted by teacher 2, and let $c_{ij}$ denote its pseudo-label
at pixel $(i,j)$. Let $\tilde P=f_{\theta_1}(\tilde X)$ denote the class
probabilities predicted by student 1. The formulation for the opposite
direction is obtained by swapping the network indices.

For teacher $s$, let $\hat b_{k,s}(j)$ denote the predicted row coordinate of
the $k$-th interface in column $j$. Let $\mathcal R_k(j)$ be the set of rows
within $\delta$ pixels of teacher 2's predicted interface $\hat b_{k,2}(j)$ in
that column. We consider the interface reliable if three conditions hold.
First, both teachers predict the interface in column $j$, and their positions
agree within $\delta$ pixels, $|\hat b_{k,2}(j)-\hat b_{k,1}(j)|\le\delta$.
Second, teacher 2's mean confidence over $\mathcal R_k(j)$ is at least
$p_{\min}$, where its confidence at each pixel is the probability
$Q_{ij,c_{ij}}$ it assigns to its own pseudo-label. Third, the pseudo-labels of
teacher 2 are nondecreasing with depth throughout column $j$.

For a reliable interface, we also measure student agreement by averaging the
probability that student 1 assigns to each of those same pseudo-labels.
Following dynamic mutual training \cite{feng2022dmt}, we compute the interface
weight as
\[
  w_{\mathrm{int}}(k,j)
  =
  \left[
  \frac{1}{|\mathcal R_k(j)|}
  \sum_{i\in\mathcal R_k(j)}
  \tilde P_{ij,c_{ij}}
  \right]^\gamma,
\]
where $\gamma>0$ controls how strongly the interface weight depends on student
agreement. We set $w_{\mathrm{int}}(k,j)=0$ if any of the three reliability
conditions fails. We do not propagate gradients through $w_{\mathrm{int}}(k,j)$. The weight is recomputed at
every training step and serves as a fixed scaling factor during each update.

We use the interface weights to derive a pixelwise weight
$w_{\mathrm{pseudo}}(i,j)$ for each pseudo-label. For an interior class
$1\le c_{ij}\le 8$, the two bounding interfaces have indices $c_{ij}$ and
$c_{ij}+1$. We assign the pixel the smaller of their weights,
\[
  w_{\mathrm{pseudo}}(i,j)
  =
  \min\!\left\{
  w_{\mathrm{int}}(c_{ij},j),\,
  w_{\mathrm{int}}(c_{ij}+1,j)
  \right\}.
\]
Classes 0 and 9 each have one adjacent interface, so their pixels receive
weights $w_{\mathrm{int}}(1,j)$ and $w_{\mathrm{int}}(9,j)$, respectively. These pixel weights determine how much each pseudo-label contributes to the unsupervised loss. Pseudo-labels bounded by an unreliable interface are ignored, and the others are weighted by student agreement.

\subsection{Training Objectives}

\subsubsection{Supervised Segmentation Loss}
Given a prediction $P$ and the class bounds,
the supervised segmentation loss for a single view combines three terms.

\paragraph{Cross-Entropy Loss.}  The
first is a cross-entropy over the admissible classes,
\[
  \mathcal L_{\mathrm{CE}}(P)
  =
  -\frac{1}{\sum_{(i,j)\in\Omega}w_{\mathrm{CE}}(i,j)}
  \sum_{(i,j)\in\Omega}w_{\mathrm{CE}}(i,j)\,
  \log\!\left(\sum_{c=L_{ij}}^{U_{ij}}P_{ijc}\right),
\]
where $w_{\mathrm{CE}}(i,j)$ is the mean class weight over the admissible
classes of the pixel. This loss term accommodates partial labels by rewarding
probability mass allocated anywhere inside the admissible interval. At a pixel
with an exact label, the interval collapses to a single class, and the negative
log term reduces to $-\log P_{ij,Y_{ij}}$.

\paragraph{Dice Loss.} The second term, $\mathcal L_{\mathrm{Dice}}(P)$, is a
soft Dice loss \cite{milletari2016vnet}. Since a pixel with several admissible
classes does not belong to a single region, we compute it only over the pixels
with an exact label, $L_{ij}=U_{ij}$.

\paragraph{Boundary Loss.} Since region-based losses such as the Dice loss are
relatively insensitive to fine boundary localization, we also incorporate a
boundary loss penalizing probability mass according to its distance from where
each class may occur. For class $c$, let $\phi_{ijc}$ be the signed distance
from pixel $(i,j)$ to the spatial bounds of its admissible layer, normalized by
the image height $H$. This distance is negative inside the region and positive
outside. Following the boundary loss \cite{kervadec2021boundary}, we define
\[
  \mathcal L_{\mathrm{bnd}}(P)
  =
  \frac{1}{|\Omega|}
  \sum_{(i,j)\in\Omega}
  \left[
  \sum_c P_{ijc}\,\phi_{ijc}
  -
  \min_{L_{ij}\le c\le U_{ij}}\phi_{ijc}
  \right].
\]
The second term in the brackets, $\min_{L_{ij}\le c\le U_{ij}}\phi_{ijc}$, does
not depend on the prediction and ensures the loss remains nonnegative. The
supervised segmentation loss can be written as
\[
  \mathcal L_{\mathrm{seg}}(P)
  =
  \mathcal L_{\mathrm{CE}}(P)
  +
  \mathcal L_{\mathrm{Dice}}(P)
  +
  \lambda_{\mathrm{bnd}}\,\mathcal L_{\mathrm{bnd}}(P),
\]
where $\lambda_{\mathrm{bnd}}$ weights the boundary term.

For a labeled patch $X$ and its perturbed view $\tilde X$, let $P=f_\theta(X)$ and
$\tilde P=f_\theta(\tilde X)$. We supervise both views and encourage their
predictions to agree,
\[
  \mathcal L_{\mathrm{sup}}
  =
  \mathcal L_{\mathrm{seg}}(P)
  +
  \lambda_{\mathrm{pert}}\,\mathcal L_{\mathrm{seg}}(\tilde P)
  +
  \lambda_{\mathrm{KL}}\,\mathcal L_{\mathrm{KL}}.
\]
Here, $\mathcal L_{\mathrm{KL}}$ is the Kullback--Leibler divergence
$\sum_c\tilde P_{ijc}\log(\tilde P_{ijc}/P_{ijc})$ averaged over all pixels of
the patch. We do not propagate gradients through $P$ in this term. The
unperturbed prediction thus serves as a fixed target for the perturbed
prediction during each update.

\paragraph{Interval share penalty.} For the SAM-based network, we add to
$\mathcal L_{\mathrm{sup}}$ a penalty that prevents a single class from
occupying the entire unannotated interval. It applies to pixels that have at
least four admissible classes. In each column $j$ that contains such pixels,
we renormalize their predicted probabilities over the admissible classes and
average them over these pixels, which gives the share $z_c(j)$ of class $c$.
Denoting the set of these columns by $\mathcal J$, the penalty is
\[
  \mathcal L_{\mathrm{ISP}}
  =
  \frac{1}{|\mathcal J|}
  \sum_{j\in\mathcal J}\sum_{c}
  \max\!\left\{0,\,z_c(j)-z_{\max}\right\}^2,
\]
where $z_{\max}$ is the largest share that is not penalized.

\subsubsection{Semi-Supervised Loss}
For an unlabeled patch, student 1 is trained on the pseudo-labels $c_{ij}$ of
teacher 2 using the pixel weights $w_{\mathrm{pseudo}}(i,j)$ defined in the
cross-teaching section. When the sum of the pixel weights is positive, the
unsupervised loss is
\[
  \mathcal L_{\mathrm{unsup}}
  =
  -\frac{1}{\sum_{i,j}w_{\mathrm{pseudo}}(i,j)}
  \sum_{i,j}w_{\mathrm{pseudo}}(i,j)\,
  \log\tilde P_{ij,c_{ij}}.
\]
The loss for student 2 is obtained by exchanging the network indices.

\paragraph{Overall objective.} In the supervised stage, each network is trained
with $\mathcal L_{\mathrm{sup}}$. In the semi-supervised stage, each student is
trained with
\[
  \mathcal L
  =
  \mathcal L_{\mathrm{sup}}
  +
  \lambda_{\mathrm{unsup}}\,\mathcal L_{\mathrm{unsup}},
\]
where $\lambda_{\mathrm{unsup}}$ weights the unsupervised term. The two students are
optimized separately at every step, and each teacher is then updated from its
student. The active loss terms and their coefficients at each stage are given
in Implementation Details.

\subsection{Post-Processing}
During inference, we reconstruct the dense probability map for the entire
B-scan using a sliding-window strategy. Specifically, each network processes
overlapping patches at the original image resolution. To smoothly blend these
predictions and suppress edge artifacts, we aggregate the patch-wise
probabilities using a spatial Gaussian window centered on each patch. This
results in the probability maps $P_{\mathrm{SAM}}$ and
$P_{\mathrm{UNet}}$ for the SAM-based network and the U-Net, respectively.

\paragraph{Fusion.} We combine the predictions of the two networks through a
weighted average,
\[
  P_{\mathrm{fuse}}
  =
  w_{\mathrm{SAM}}\,P_{\mathrm{SAM}}
  +
  (1-w_{\mathrm{SAM}})\,P_{\mathrm{UNet}},
\]
where $w_{\mathrm{SAM}}\in[0,1]$ is the weight assigned to the SAM-based
network. If exactly one network predicts implausible layer thicknesses across a
large fraction of columns, we shift the weight toward the more anatomically
plausible prediction.

\paragraph{Viterbi Decoding.} Raw pixel-wise predictions can violate the
monotonic depth order of retinal layers. To enforce this order, we decode all
pixels in a column jointly. We restrict the search to classes present in the
initial predictions and discard any retinal class predicted in very few pixels
with low confidence. For each column, we seek the label sequence that
maximizes the total log-probability assigned by $P_{\mathrm{fuse}}$ under the
constraint that the classes are nondecreasing with depth. We solve this 1D
optimization exactly using the Viterbi algorithm \cite{viterbi1967error,forney1973viterbi}.

\section{Evaluation}

\subsection{Datasets}

We train \SoftwareName{} on both labeled and unlabeled OCT B-scans (Table~\ref{tab:data}). Since different datasets follow different annotation protocols, we map their annotations to the nine interfaces.

\begin{table}[htbp]
  \centering
  \begin{threeparttable}
  \caption{The Nine Retinal Interfaces. A slash denotes the boundary between two retinal layers.}
  \label{tab:interfaces}
  \small
  \begin{tabular}{@{}lll@{}}
    \toprule
    Interface & Abbreviation & Name \\
    \midrule
    $b_1$ & ILM & Inner limiting membrane \\
    $b_2$ & RNFL/GCL & Retinal nerve fiber layer / ganglion cell layer \\
    $b_3$ & GCL/IPL & Ganglion cell layer / inner plexiform layer \\
    $b_4$ & IPL/INL & Inner plexiform layer / inner nuclear layer \\
    $b_5$ & INL/OPL & Inner nuclear layer / outer plexiform layer \\
    $b_6$ & ELM & External limiting membrane \\
    $b_7$ & IS/OS & Photoreceptor inner segment / outer segment (ellipsoid zone) \\
    $b_8$ & OS/RPE & Photoreceptor outer segment / retinal pigment epithelium \\
    $b_9$ & BM & Bruch's membrane \\
    \bottomrule
  \end{tabular}
  \end{threeparttable}
\end{table}

\begin{table}[htbp]
  \centering
  \begin{threeparttable}
  \caption{Training Data used in \SoftwareName{}. The nine annotated interfaces are indicated by $\bullet$.}
  \label{tab:data}
  \small
  \setlength{\tabcolsep}{4pt}
  \newcommand{\an}{$\bullet$}%
  \begin{tabularx}{\linewidth}{@{}>{\raggedright\arraybackslash}Xlrr@{\hskip 14pt}*{9}{c}@{\hskip 14pt}l@{}}
    \toprule
    & & \multicolumn{2}{c}{B-scans} & \multicolumn{9}{c}{Annotated interfaces $b_k$} & \\
    \cmidrule(lr){3-4}\cmidrule(lr){5-13}
    Dataset & Device & Labeled & Unlabeled & 1 & 2 & 3 & 4 & 5 & 6 & 7 & 8 & 9 & Label source \\
    \midrule
    Synthetic \cite{owen2026daoct} & -- & 186 & -- & \an & & & & \an & & \an & \an & \an & Synthetic \\
    JHU HC-MS \cite{he2019jhu} & Spectralis & 1,715 & -- & \an & \an & & \an & \an & \an & \an & \an & \an & Manual \\
    Duke DME \cite{chiu2015dme} & Spectralis & 110 & -- & \an & \an & & \an & \an & & \an & \an & \an & Manual \\
    OCT5k \cite{arikan2025oct5k} & Spectralis & 1,672 & -- & \an & & & & \an & & \an & \an & \an & Manual \\
    AI-READI \cite{aireadi2025flagship} & Spectralis & 26,028 & 103,680 & \an & \an & \an & \an & \an & \an & \an & \an & \an & Device software \\
    & Maestro2 & 52,597 & 107,143 & \an & \an & \an & \an & \an & \an & \an & \an & \an & Device software \\
    & Triton & 103,623 & 53,883 & \an & \an & \an & \an & \an & \an & \an & \an & \an & Device software \\
    & Cirrus & 65,475 & 112,940 & \an & & & & & & & & \an & Device software \\
    \midrule
    Total & & 251,406 & 377,646 & \multicolumn{9}{c}{} & \\
    \bottomrule
  \end{tabularx}
  \end{threeparttable}
\end{table}

\paragraph{Synthetic datasets.} The DA-OCT challenge \cite{owen2026daoct} released
230 synthetic B-scans with pixelwise label maps. We use five
interfaces that match the correct target interfaces. 186 of these images are used for training and the rest are reserved for local evaluation.

\paragraph{Public datasets.} Three public datasets provide manual annotations of scans acquired with the Spectralis (Heidelberg Engineering, Heidelberg, Germany). The JHU HC-MS dataset \cite{he2019jhu} includes healthy controls and patients with multiple sclerosis. The Duke DME dataset \cite{chiu2015dme} contains scans from patients with diabetic macular edema. We also include the OCT5k dataset \cite{arikan2025oct5k}. Since Duke DME and OCT5k were annotated by more than one grader, we use the annotations from the first grader.

\paragraph{AI-READI.} Most of our training data comes from AI-READI \cite{aireadi2024rethinking,aireadi2025flagship,owsley2025protocol}, which comprises 2,280 participants and includes OCT volumes acquired
with the Spectralis, the Maestro2 and the Triton (Topcon, Tokyo, Japan), and
the Cirrus (Carl Zeiss Meditec, Dublin, CA, USA). We use the segmentations
generated by the device software as labels in place of manual annotations. For
Cirrus, they contain only the inner limiting membrane ($b_1$) and the center
of the retinal pigment epithelium, which we shift 9 pixels deeper to
approximate Bruch's membrane ($b_9$). We sample up to 16 B-scans from each
volume, and at most four from each Cirrus optic disc volume, which results in
247,723 labeled B-scans from 1,685 participants. The unlabeled B-scans come
from the same participants.

\subsubsection{Test sets} The DA-OCT challenge is built on AI-READI with hidden validation and test sets manually annotated by retina specialists \cite{owen2026daoct}. Each submission is first trained on the challenge server, where only the macular B-scans from the Maestro2 are labeled and the Spectralis, Cirrus, and widefield Maestro2 B-scans are unlabeled. It is then evaluated on 628 B-scans in the validation phase and 658 in the final phase. 
The submissions are ranked on macular and widefield B-scans from all four devices.
Since the images and annotations remain on the challenge server and are not part of the public AI-READI release, we cannot use these test sets in our other experiments.

To evaluate our methods, we utilize the local test sets built from AI-READI participants who are excluded from the training data. The first set consists of 195 Spectralis B-scans from 40 participants. Its labels are generated by the device software, and we refer to it as Machine Labeled in the table. The second set consists of 124 macular B-scans from 96 participants acquired with all four devices and 49 widefield B-scans from 36 participants acquired with the Maestro2. These B-scans were manually annotated by the first author, and we refer to this set
as Manual Labeled. We note here that these annotations are not expert annotations and have no clinical reference value. We use them only to compare different methods on the same B-scans.
We also include an external test set from the Duke AMD dataset \cite{chiu2012amd}. It provides manual annotations of three interfaces for 220 B-scans from 20 volumes of patients with age-related macular degeneration, acquired with a
Bioptigen SD-OCT system (Bioptigen, Research Triangle Park, NC, USA). This dataset is not used for training, and its device does not appear in our training data. Since it was annotated by two graders, we use the annotations from the first grader.

On both the Machine Labeled set and the Manual Labeled set, we exclude $b_2$ and $b_3$ because we believe that their correct anatomical positions cannot be clearly identified. The Manual Labeled annotations also do not contain $b_4$. The external test set only provides three interfaces ($b_1$, $b_8$, and $b_9$).

\subsection{Evaluation Metrics}
We evaluate the segmentations with the Dice score and the mean absolute error (MAE) of the interface positions. For the MAE, we first obtain the interface positions from the predicted label map. In each column, the position of $b_k$ is the first row whose class is at least $k$. We then measure the absolute distance in pixels between the predicted and the annotated positions. 

In the ablation studies, we report the image-level score of the challenge
\cite{owen2026daoct} on both the Machine Labeled set and the Manual Labeled
set. For each class, the score of a B-scan is computed as
\[
  \frac{1}{2}\,\mathrm{Dice}
  +
  \frac{1}{2}\exp\!\left[-\frac{\mathrm{MASD}/H}{0.02}\right],
\]
where MASD is the mean average surface distance
of the class, and $H$ is the image height. Each class occupies a region of the B-scan. MASD measures how far the outline of this region in the prediction lies from its outline in the annotation. For a retinal layer, the outline
consists of the two interfaces that bound it. We average this score over the classes that are not bounded by an excluded interface and then over the
B-scans.

\subsection{Comparison With Existing Methods}
We compare \SoftwareName{} with nine segmentation models (Table~\ref{tab:comparison}). The CNN-based model includes U-Net
\cite{ronneberger2015unet}, ReLayNet \cite{roy2017relaynet}, MGU-Net
\cite{li2021peripapillary}, and nnU-Net 
\cite{isensee2021nnunet}. The Transformer-based model includes TransUNet \cite{chen2021transunet} and Swin-UNETR \cite{hatamizadeh2022swinunetr}. The foundation model includes SAM \cite{kirillov2023sam}, RETFound \cite{zhou2023retfound}, and MIRAGE \cite{morano2025mirage}. We train all baselines on the same labeled B-scans as our U-Net. All baselines share the same number of training updates, standard augmentations, supervised segmentation loss $\mathcal L_{\mathrm{seg}}(P)$,
sliding-window inference, and Viterbi decoding. For each foundation model, we attach the decoder of our SAM-based network to its
pre-trained encoder. As for our SAM-based network, we first train only the
decoder and then fine-tune all encoder blocks with a layer-wise learning rate
decay. The SAM and nnU-Net baselines therefore have the same architectures as
the two networks of \SoftwareName{}. We train every method with three random seeds and report the mean and standard deviation.

\begin{table}[htbp]
\centering
\caption{Comparison With Existing Methods. Values are the mean $\pm$ standard deviation over three seeds. The Manual Labeled columns report the widefield B-scans. The best results are in bold. The second best results are underlined.}\label{tab:comparison}
\providecolor{octnavy}{HTML}{2E4775}
\setlength{\aboverulesep}{0pt}\setlength{\belowrulesep}{0pt}\renewcommand{\arraystretch}{1.18}
\resizebox{\ifdim\resulttablescale\width>\linewidth\linewidth\else\resulttablescale\width\fi}{!}{%
\begin{tabular}{llrrrrrrr}
\toprule
 &  &  & \multicolumn{4}{c}{AI-READI} & \multicolumn{2}{c}{External} \\
\cmidrule(lr){4-7}\cmidrule(lr){8-9}
 &  &  & \multicolumn{2}{c}{Machine Labeled} & \multicolumn{2}{c}{Manual Labeled} &  &  \\
\cmidrule(lr){4-5}\cmidrule(lr){6-7}
 & Method & Params (M) & Dice $\uparrow$ & MAE $\downarrow$ & Dice $\uparrow$ & MAE $\downarrow$ & Dice $\uparrow$ & MAE $\downarrow$ \\
\midrule
 & U-Net & 31.0 & \textbf{92.9}\,{\scriptsize$\pm$\,0.05} & 0.78\,{\scriptsize$\pm$\,0.021} & 73.8\,{\scriptsize$\pm$\,1.05} & 6.37\,{\scriptsize$\pm$\,0.902} & 77.5\,{\scriptsize$\pm$\,2.43} & 5.70\,{\scriptsize$\pm$\,0.968} \\
 & ReLayNet & 0.8 & 91.5\,{\scriptsize$\pm$\,0.06} & 0.95\,{\scriptsize$\pm$\,0.019} & 75.5\,{\scriptsize$\pm$\,0.35} & 4.26\,{\scriptsize$\pm$\,0.048} & 69.1\,{\scriptsize$\pm$\,2.47} & 8.87\,{\scriptsize$\pm$\,1.421} \\
 & MGU-Net & 2.1 & 92.1\,{\scriptsize$\pm$\,0.08} & 0.85\,{\scriptsize$\pm$\,0.015} & 74.2\,{\scriptsize$\pm$\,1.30} & 5.61\,{\scriptsize$\pm$\,0.504} & 82.5\,{\scriptsize$\pm$\,2.15} & 4.02\,{\scriptsize$\pm$\,0.836} \\
\multirow{-4}{*}{CNN} & nnU-Net & 33.5 & \underline{92.8}\,{\scriptsize$\pm$\,0.13} & 0.78\,{\scriptsize$\pm$\,0.032} & 73.8\,{\scriptsize$\pm$\,1.32} & 5.70\,{\scriptsize$\pm$\,1.238} & 82.5\,{\scriptsize$\pm$\,1.40} & 2.94\,{\scriptsize$\pm$\,0.124} \\
\specialrule{0.4pt}{0pt}{0pt}
 & TransUNet & 105.3 & \underline{92.8}\,{\scriptsize$\pm$\,0.11} & \underline{0.77}\,{\scriptsize$\pm$\,0.015} & 65.6\,{\scriptsize$\pm$\,3.94} & 10.00\,{\scriptsize$\pm$\,1.852} & 83.1\,{\scriptsize$\pm$\,0.80} & 3.74\,{\scriptsize$\pm$\,0.992} \\
\multirow{-2}{*}{Transformer} & Swin-UNETR & 25.1 & 92.5\,{\scriptsize$\pm$\,0.05} & 0.86\,{\scriptsize$\pm$\,0.006} & \underline{76.5}\,{\scriptsize$\pm$\,0.66} & \textbf{4.13}\,{\scriptsize$\pm$\,0.149} & 82.6\,{\scriptsize$\pm$\,1.79} & 3.37\,{\scriptsize$\pm$\,0.164} \\
\specialrule{0.4pt}{0pt}{0pt}
 & SAM & 313.1 & 92.6\,{\scriptsize$\pm$\,0.06} & 0.78\,{\scriptsize$\pm$\,0.004} & 51.1\,{\scriptsize$\pm$\,4.24} & 25.87\,{\scriptsize$\pm$\,2.995} & \textbf{84.3}\,{\scriptsize$\pm$\,0.36} & \underline{2.81}\,{\scriptsize$\pm$\,0.041} \\
 & RETFound & 308.9 & 92.3\,{\scriptsize$\pm$\,0.00} & 0.83\,{\scriptsize$\pm$\,0.007} & 55.8\,{\scriptsize$\pm$\,1.00} & 21.80\,{\scriptsize$\pm$\,0.976} & 83.3\,{\scriptsize$\pm$\,0.31} & 3.34\,{\scriptsize$\pm$\,0.220} \\
\multirow{-3}{*}{Foundation} & MIRAGE & 308.9 & 91.8\,{\scriptsize$\pm$\,0.06} & 0.89\,{\scriptsize$\pm$\,0.023} & 54.0\,{\scriptsize$\pm$\,1.69} & 21.17\,{\scriptsize$\pm$\,1.793} & 81.8\,{\scriptsize$\pm$\,0.34} & 3.46\,{\scriptsize$\pm$\,0.082} \\
\specialrule{0.8pt}{0pt}{0pt}
Ours & \cellcolor{octnavy!12}Full method & \cellcolor{octnavy!12}346.5 & \cellcolor{octnavy!12}\textbf{92.9}\,{\scriptsize$\pm$\,0.04} & \cellcolor{octnavy!12}\textbf{0.76}\,{\scriptsize$\pm$\,0.004} & \cellcolor{octnavy!12}\textbf{77.5}\,{\scriptsize$\pm$\,0.58} & \cellcolor{octnavy!12}\underline{4.15}\,{\scriptsize$\pm$\,0.370} & \cellcolor{octnavy!12}\underline{84.0}\,{\scriptsize$\pm$\,0.04} & \cellcolor{octnavy!12}\textbf{2.49}\,{\scriptsize$\pm$\,0.039} \\
\bottomrule
\end{tabular}}
\end{table}

The device and label source of the Machine Labeled set were both represented in the training data, with test participants excluded from training. On this set, all methods in Table~\ref{tab:comparison} performed similarly, with Dice scores ranging from 91.5\% to 92.9\%. On the widefield subset of the Manual Labeled set, \SoftwareName{} achieved the highest Dice score of 77.5\%, while its MAE of 4.15 pixels was comparable to the lowest MAE of 4.13 pixels achieved by Swin-UNETR. The SAM and nnU-Net baselines, corresponding to the two constituent architectures used in \SoftwareName{}, achieved Dice scores of 51.1\% and 73.8\%, respectively. On the external test set acquired with an unseen Bioptigen device, \SoftwareName{} achieved the lowest MAE of 2.49 pixels and a Dice score of 84.0\%, compared with the highest Dice score of 84.3\% achieved by SAM.

\begin{figure}[htbp]
  \centering
  \includegraphics[width=\linewidth,
    alt={Segmentations of three B-scans, showing the fovea, macular edema,
    and the optic disc, by four baseline methods and by the proposed
    software}]{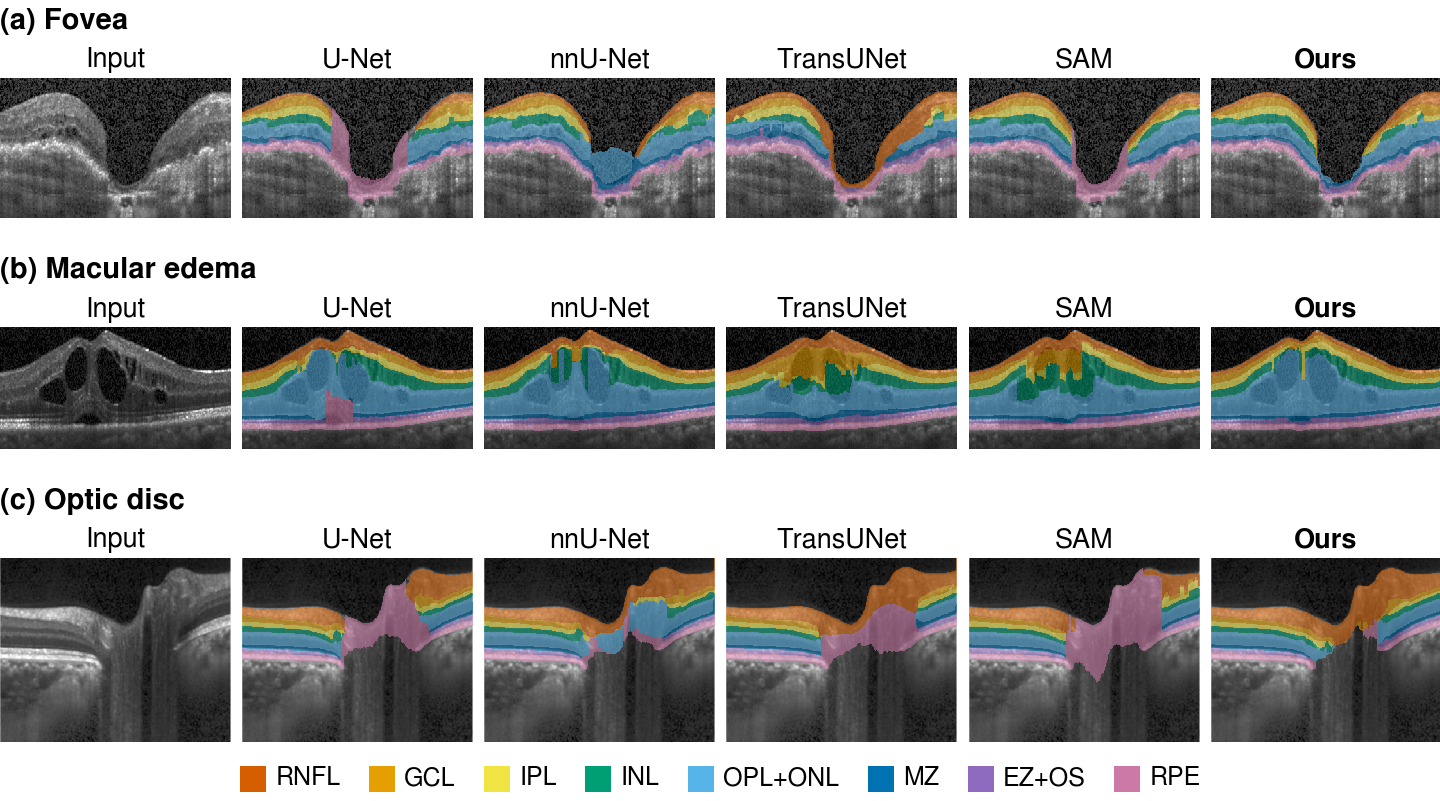}
  \caption{Example outputs of \SoftwareName{} (Ours) and four baseline methods
    for B-scans showing (a) the fovea, (b) macular edema, and (c) the optic
    disc.}
  \label{fig:outputs}
\end{figure}

Figure~\ref{fig:outputs} shows representative outputs of \SoftwareName{} and four baseline methods for B-scans containing the fovea, macular edema, and optic disc. In these examples, several baseline methods produced anatomically implausible layer segmentations around these challenging structures, whereas \SoftwareName{} better preserved anatomically plausible layer organization. These examples illustrate the types of anatomical variation that motivated our anatomy-aware design.

\subsection{Ablation Studies}
We conducted ablation studies to quantify the contributions of the key components of \SoftwareName{} (Table~\ref{tab:ablation}). The first row fuses the two networks after a supervised stage using standard augmentations only. The last row gives the full framework. On the Manual Labeled set, we report three subsets separately: the Triton 12$\times$12 mm B-scans, the remaining macular B-scans, and the widefield B-scans. For the remaining macular B-scans, we average the scores of the Spectralis, Maestro2, Cirrus, and Triton 6$\times$6 mm scans with equal weight
for each device. 

\begin{table}[htbp]
\centering
\caption{Ablation of the Components of \SoftwareName{}. The best results are in bold. The second best results are underlined.}\label{tab:ablation}
\providecolor{octnavy}{HTML}{2E4775}
\setlength{\aboverulesep}{0pt}\setlength{\belowrulesep}{0pt}\renewcommand{\arraystretch}{1.18}
\resizebox{\ifdim\resulttablescale\width>\linewidth\linewidth\else\resulttablescale\width\fi}{!}{%
\begin{tabular}{cccrrrrrr}
\toprule
 &  &  & \multicolumn{4}{c}{AI-READI} & \multicolumn{2}{c}{External} \\
\cmidrule(lr){4-7}\cmidrule(lr){8-9}
 &  &  & \multicolumn{3}{c}{Manual Labeled} & \multicolumn{1}{c}{Machine Labeled} &  &  \\
\cmidrule(lr){4-6}\cmidrule(lr){7-7}
\begin{tabular}[b]{@{}c@{}}Proposed\\augmentation\end{tabular} & \begin{tabular}[b]{@{}c@{}}Cross-\\teaching\end{tabular} & \begin{tabular}[b]{@{}c@{}}Post-\\processing\end{tabular} & Macula $\uparrow$ & Triton 12$\times$12 $\uparrow$ & Widefield $\uparrow$ & Spectralis $\uparrow$ & Dice $\uparrow$ & MAE $\downarrow$ \\
\midrule
 &  &  & 0.855 & 0.819 & 0.835 & 0.938 & 0.825 & 5.10 \\
\checkmark &  &  & 0.858 & \underline{0.832} & \underline{0.843} & 0.940 & \textbf{0.844} & \underline{3.13} \\
\checkmark & \checkmark &  & \underline{0.867} & 0.825 & 0.836 & \underline{0.942} & \underline{0.840} & 3.52 \\
\cellcolor{octnavy!12}\checkmark & \cellcolor{octnavy!12}\checkmark & \cellcolor{octnavy!12}\checkmark & \cellcolor{octnavy!12}\textbf{0.900} & \cellcolor{octnavy!12}\textbf{0.860} & \cellcolor{octnavy!12}\textbf{0.873} & \cellcolor{octnavy!12}\textbf{0.944} & \cellcolor{octnavy!12}\underline{0.840} & \cellcolor{octnavy!12}\textbf{2.46} \\
\bottomrule
\end{tabular}}
\end{table}

Adding the anatomy-aware augmentations and perturbations raised the score on the
Triton 12$\times$12 mm B-scans from 0.819 to 0.832. On the external test set,
it raised the Dice score from 0.825 to 0.844 and reduced the MAE from 5.10 to
3.13 pixels. The semi-supervised stage raised the macular score from 0.858 to
0.867. The other scores changed by at most 0.007, and the MAE on the external
test set increased from 3.13 to 3.52 pixels. Viterbi decoding raised all three
scores on the Manual Labeled set by more than 0.03 and reduced the MAE on the
external test set from 3.52 to 2.46 pixels. With all components, the full
framework achieved the best result in every column except the Dice score on
the external test set.

We also examine the loss terms of the supervised stage. Removing $\mathcal L_{\mathrm{Dice}}$
or $\mathcal L_{\mathrm{bnd}}$ changed each score by at most 0.005. Removing $\mathcal L_{\mathrm{seg}}(\tilde P)$ lowered the scores on all three subsets of the Manual Labeled set, and the widefield score dropped from 0.837 to 0.782.

\begin{table}[htbp]
\centering
\caption{Ablation of the Loss Terms in the Supervised Stage. A check mark indicates that the term is used. All columns report the challenge score without Viterbi decoding. The best results are in bold. The second best results are underlined.}\label{tab:loss}
\providecolor{octnavy}{HTML}{2E4775}
\setlength{\aboverulesep}{0pt}\setlength{\belowrulesep}{0pt}\renewcommand{\arraystretch}{1.18}
\resizebox{\ifdim\resulttablescale\width>\linewidth\linewidth\else\resulttablescale\width\fi}{!}{%
\begin{tabular}{cccrrrr}
\toprule
 &  &  & \multicolumn{4}{c}{AI-READI} \\
\cmidrule(lr){4-7}
 &  &  & \multicolumn{3}{c}{Manual Labeled} & \multicolumn{1}{c}{Machine Labeled} \\
\cmidrule(lr){4-6}\cmidrule(lr){7-7}
\begin{tabular}[b]{@{}c@{}}$\mathcal{L}_{\mathrm{Dice}}$\end{tabular} & \begin{tabular}[b]{@{}c@{}}$\mathcal{L}_{\mathrm{bnd}}$\end{tabular} & \begin{tabular}[b]{@{}c@{}}$\mathcal{L}_{\mathrm{seg}}(\tilde{P})$\end{tabular} & Macula $\uparrow$ & Triton 12$\times$12 $\uparrow$ & Widefield $\uparrow$ & Spectralis $\uparrow$ \\
\midrule
\checkmark & \checkmark & \checkmark & \underline{0.856} & \textbf{0.830} & \textbf{0.837} & \textbf{0.941} \\
 & \checkmark & \checkmark & \textbf{0.860} & \underline{0.825} & \underline{0.835} & \textbf{0.941} \\
\checkmark &  & \checkmark & 0.853 & \textbf{0.830} & 0.834 & \underline{0.940} \\
\checkmark & \checkmark &  & 0.839 & 0.815 & 0.782 & 0.939 \\
\bottomrule
\end{tabular}}
\end{table}

\section{Software Summary}

\SoftwareName{} was developed as an open-source framework for retinal layer
segmentation in OCT B-scans acquired across different devices. It operates on
B-scan images and outputs a label map of the scan in which every pixel is
assigned to one of ten ordered classes separated by nine retinal interfaces.
The framework implements the complete workflow behind our winning entry to the
MICCAI 2026 Domain Adaptation for OCT (DA-OCT) challenge \cite{owen2026daoct}. It was developed in
Python on top of PyTorch, using timm, NumPy, OpenCV, SciPy, and other
open-source dependencies. A MONAI bundle with both models and weights can be
obtained directly from the MONAI Model Zoo \cite{cardoso2022monai} for immediate use. Alternatively,
the complete framework and model weights can be downloaded from GitHub and
Hugging Face, respectively. Models obtained from both sources are released
under the BSD 2-Clause license. The package is designed to be run from the
command line on a local machine or as jobs on a Slurm cluster for large-scale
workflows. It provides workflows to either use the pretrained model directly or
train your own model on different datasets.

\subsection{Usage}
The trained model checkpoint from the challenge can be obtained in two ways:

\paragraph{MONAI bundle.} To run inference directly, download the MONAI bundle
that contains both the model definitions and weights with:
\begin{lstlisting}[language=bash]
pip install "monai[fire]"
python -m monai.bundle download --name daoct_retinal_seg_yzjaxon \
    --bundle_dir <BUNDLE_DIR>
\end{lstlisting}
To generate segmentation masks for input OCT scans in a folder
\texttt{<IMAGE\_DIR>} into a desired output directory \texttt{<MASK\_DIR>}, run:
\begin{lstlisting}[language=bash]
python <BUNDLE_DIR>/daoct_retinal_seg_yzjaxon/infer_bundle.py \
    --input <IMAGE_DIR> --output <MASK_DIR>
\end{lstlisting}
where \texttt{<IMAGE\_DIR>} is a folder containing B-scans in PNG format and
\texttt{<MASK\_DIR>} is the desired output directory. This command stores one
label map per input image in a folder structure with a matching filename of
\texttt{<MASK\_DIR>/<name>-mask.png}. Inference can be run on either CPU or GPU
and requires no additional input metadata.

\paragraph{GitHub repository.} For developers seeking to train new models,
adapt the framework, or reproduce our results, the full source code is
available on our GitHub page.\cite{softwarerepo}

The repository pins its dependencies and records reference values for the
software environment, the prepared datasets, and the training runs. Before
running the code, set up the environment and configure
\texttt{configs/local.yaml}, which stores the paths and execution settings
required by the program. Detailed installation instructions are provided in
the repository. We provide commands for three workflows:

\textbf{1.~Checking the installation and segmenting with the submitted
weights.}
\begin{lstlisting}[language=bash]
python scripts/repro.py check all
python scripts/repro.py infer <IMAGE_DIR> <MASK_DIR>
\end{lstlisting}
The \texttt{check} command compares the local environment and available inputs
against the recorded references and reports any differences. This design is
intended to facilitate exact reproduction of the training setup. The
\texttt{infer} command runs segmentation on the input images and writes masks
in the same format as the MONAI bundle.

\textbf{2.~Training a new model.} Each model from the framework can be trained
from its own configuration file and exported for inference individually:
\begin{lstlisting}[language=bash]
python scripts/repro.py run configs/framework/cnn.yaml
python scripts/repro.py export <RUN_DIR>/checkpoints/last.pt <EXPORT_DIR>
\end{lstlisting}
The example configuration trains a PlainConvUNet from scratch on the labeled
synthetic images released by the challenge organizers. Each run stores its
fully resolved configuration and its checkpoints in \texttt{<RUN\_DIR>}. This
training step requires a GPU.

\textbf{3.~Reproduction of the challenge models.} We provide commands to
download the publicly available training data, including the JHU HC-MS and
Duke DME datasets, the OCT5k annotations, and the official challenge data, as
well as the SAM initialization weights for fine-tuning. Note that the OCT5k
source images and the AI-READI dataset must be obtained separately from their
providers under their respective data license agreements.
\begin{lstlisting}[language=bash]
python scripts/repro.py download all
python scripts/repro.py reproduce
python scripts/repro.py score --weights mine
\end{lstlisting}
The \texttt{reproduce} command builds the training pools, executes all
training stages, and exports the model pair. The \texttt{score} command runs
inference on the 230 synthetic images released by the organizers and evaluates
the predictions with the official scoring program. Since most of these images
were used for training, this score serves only to verify pipeline
reproducibility. Adding \texttt{-{}-dry} to any command prints the planned
steps without executing them, allowing users to check the workflow beforehand.
Each training run uses a single GPU.

\subsection{Implementation Details}
We summarize the main settings to reproduce our training pipeline below. The two networks that we fuse come from two
separate trainings of the full pipeline. The first training did not use the optic-disc-like depression, whereas the second applied it in every run. We fuse the SAM-based student of the first training with the U-Net student of the second for the final submission models to the challenge. This combination performed best on our local test sets. Every run uses a single NVIDIA H200 GPU,
and the two trainings of the full pipeline together took 14.5 GPU hours. The fusion weight $w_{\mathrm{SAM}}$ is 0.35 for macular scans and 0.5 for widefield scans.

The SAM-based network $f_{\theta_1}$ combines the
ViT-L/16 image encoder of SAM with a decoder that builds a feature pyramid from intermediate encoder features, as in DPT \cite{ranftl2021dpt}. The U-Net $f_{\theta_2}$ follows the nnU-Net design \cite{isensee2021nnunet}. Both are trained with deep supervision. We clip the intensities of each
B-scan to its 0.5th and 99.5th percentiles and normalize them to zero mean and unit variance. During training, we apply the anatomy-aware augmentations to the whole normalized B-scan, crop a patch of $256\times256$ pixels, and then apply standard spatial and intensity augmentations. Local morphological deformation and the optic-disc-like depression are applied with probabilities
of 0.3 and 0.25, respectively, and a perturbed view is generated for every patch.

All runs use AdamW \cite{loshchilov2019adamw} with a batch size
of 8. We train the SAM-based network in two runs. The first trains only the decoder with $\mathcal L_{\mathrm{seg}}(P)$ on the unperturbed patch, and the second fine-tunes all encoder blocks with a layer-wise learning rate decay. We
set $\lambda_{\mathrm{pert}}=\lambda_{\mathrm{KL}}=0.5$ and $z_{\max}=0.6$. After a linear warm-up in each run, $\lambda_{\mathrm{bnd}}$ is 0.2 for the SAM-based network and 0.5 for the U-Net. The class weights that enter
$w_{\mathrm{CE}}$ are inversely proportional to the square root of the class frequencies in the synthetic training images.

Cross-teaching starts from the networks of the supervised stage and runs for a
single pass over the unlabeled data, with $\delta=3$ pixels, $p_{\min}=0.8$,
$\gamma=3$, and $\lambda_{\mathrm{unsup}}=0.5$. On labeled
patches, the SAM-based student is trained with
$\mathcal L_{\mathrm{sup}}+\mathcal L_{\mathrm{ISP}}$, while the U-Net student is trained with $\mathcal L_{\mathrm{seg}}(P)$ on the unperturbed patch only.

\begin{figure}[htbp]
  \centering
  \includegraphics[width=0.6\linewidth,
    alt={Scatter plot of the macular and widefield scores of ten challenge
    teams}]{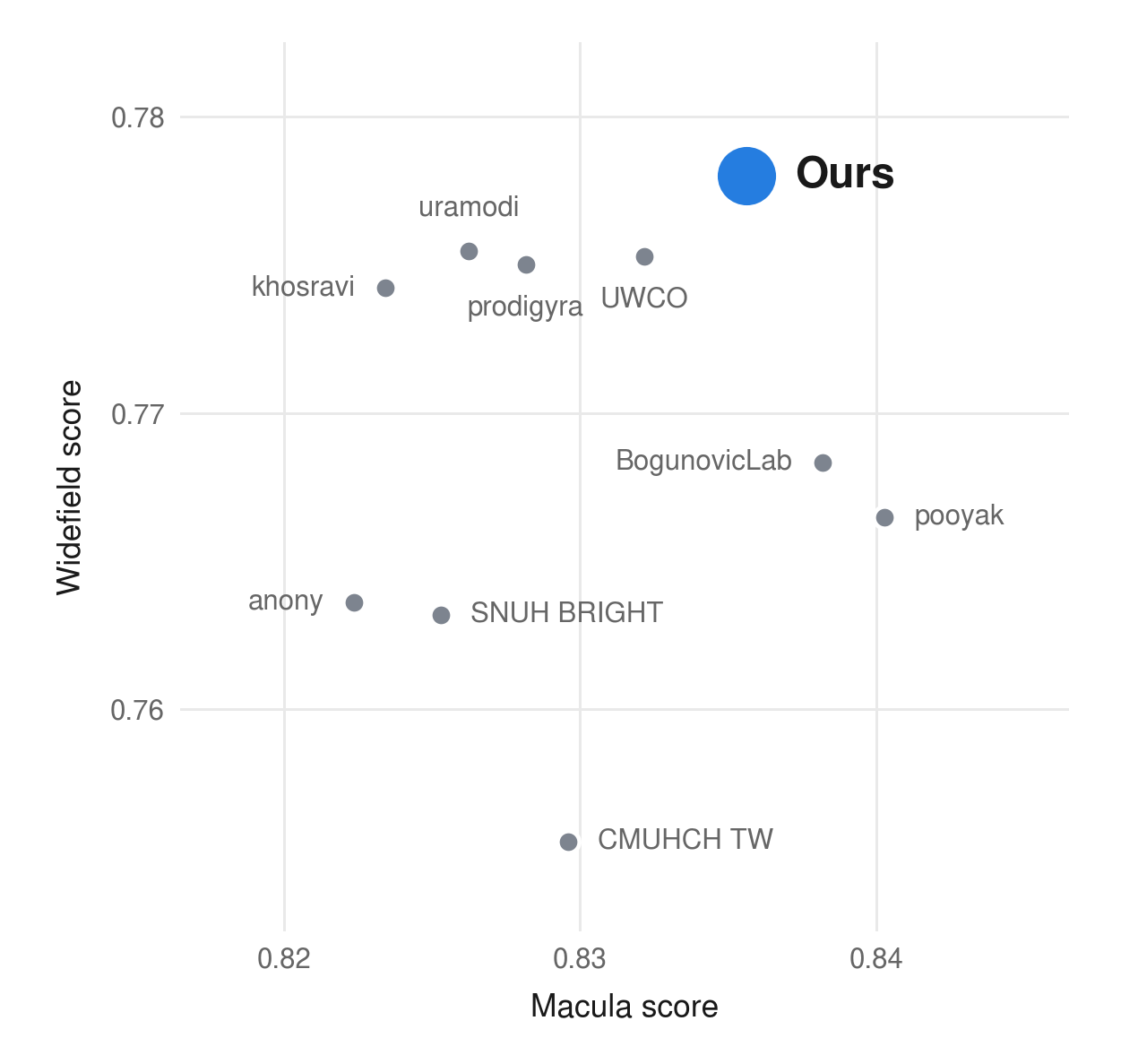}
  \caption{Macular and widefield scores on the hidden test set of the DA-OCT
    challenge. Each point represents one team. The point labeled ``Ours'' is
    \SoftwareName{}.}
  \label{fig:scores}
\end{figure}

\paragraph{Challenge submission.} For the challenge, the two fused networks
were further adapted on the evaluation server with a second round of
cross-teaching on 513 annotated Maestro2 B-scans and 4,367 unlabeled B-scans
from the target domains. The adapted system scored 0.8068 on the hidden test
set and ranked first. Figure~\ref{fig:scores} shows its macular and widefield scores together with those of other teams.

\section{Conclusions}

Overall, \SoftwareName{} integrates anatomy-aware augmentation, semi-supervised cross-teaching, and anatomically constrained post-processing for cross-device retinal OCT segmentation. The method ranked first in the MICCAI 2026 DA-OCT Challenge and demonstrated robust performance across multiple OCT devices and an external dataset. The complete framework, trained models, and reproducibility workflows are openly available to facilitate further research and application.

\section*{Acknowledgments}

The authors thank the organizers of the MICCAI 2026 Domain Adaptation for OCT Segmentation (DA-OCT) Challenge for organizing and supporting the challenge. We also thank the clinical experts for their efforts in annotating the challenge data and the AI-READI Consortium and its contributors for developing and providing the dataset that supported this work. We thank Younjoon Chung for helpful suggestions on the methods of this work. This work was supported in part by the National Library of Medicine of the National Institutes of Health under Award No. 4R00LM014024.

Google Gemini 4 Hydrogen was used for proofreading and refining the writing of this manuscript in November 2026.

\bibliographystyle{tvst}
\bibliography{references}

\end{document}